\relax
\documentclass[letterpaper]{article} 
\usepackage{aaai21} 
\usepackage{times} 
\usepackage{helvet} 
\usepackage{courier} 
\usepackage[hyphens]{url} 
\usepackage{graphicx} 
\usepackage{graphicx} 
\usepackage{natbib} 
\usepackage{caption} 
\usepackage{amsmath}
\usepackage{amssymb}
\usepackage{multirow}
\title{I-GCN: Robust Graph Convolutional Network via Influence Mechanism }
\author {

        Haoxi Zhan,\textsuperscript{\rm 1}
        Xiaobing Pei\thanks{Corresponding Author.} \textsuperscript{\rm 1} \\
}
\affiliations {
    \textsuperscript{\rm 1} Huazhong University of Science and Technology \\
   zhanhaoxi@foxmail.com , xiaobingp@hust.edu.cn 
}

\begin{document}

\maketitle

\begin{abstract}

Deep learning models for graphs, especially Graph Convolutional Networks (GCNs), have achieved remarkable performance in the task of semi-supervised node classification. However, recent studies show that GCNs suffer from adversarial perturbations. Such vulnerability to adversarial attacks significantly decreases the stability of GCNs when being applied to security-critical applications. Defense methods such as preprocessing, attention mechanism and adversarial training have been discussed by various studies. While being able to achieve desirable performance when the perturbation rates are low, such methods are still vulnerable to high perturbation rates. Meanwhile, some defending algorithms perform poorly when the node features are not visible. Therefore, in this paper, we propose a novel mechanism called influence mechanism, which is able to enhance the robustness of the GCNs significantly. The influence mechanism divides the effect of each node into two parts: introverted influence which tries to maintain its own features and extroverted influence which exerts influences on other nodes. Utilizing the influence mechanism, we propose the Influence GCN (I-GCN) model. Extensive experiments show that our proposed model is able to achieve higher accuracy rates than state-of-the-art methods when defending against non-targeted attacks.

\end{abstract}

\section{Introduction}

Graphs are ubiquitous discrete data structures which consist of nodes and edges. A number of complex relationships in non-Euclidean domains such as social networks\cite{Social}, citation networks\cite{Citation}, bank transactions\cite{Transaction} and protein interaction networks\cite{Protein} are able to be represented by graphs. Not surprisingly, recent years have witnessed a rapid growth in the applications of Graph Neural Networks (GNNs). One important analytical task of GNNs is node classification, which aims to predict the labels of the unlabeled nodes in a graph. Graph Convolutional Networks (GCNs) are a type of GNNs designed to tackle the node classification task that attracts considerable research attention\cite{Kipf, GAT, GraphSAGE}. 

Although the performance of GCNs are promising, deep learning methods are shown to be in lack of robustness\cite{AdvI}. Unfortunately, GCNs are of no exception. It is revealed that only a small portion of unnoticeable perturbations would be able to fool Graph Convolutional Networks such that the classification accuracies would be highly reduced\cite{Nettack, Mettack}. The lack of robustness is one of the major obstacles for GCNs to be applied in the real world. For instance, in an advertisement system based on social network graphs, an attacker might create several fake edges in the data to mislead the user classification system. Similarly, adding adversarial transactions in a bank system could lead to false detections of fraud users. Therefore, it is significant to develop GCN models that are robust against such adversarial attacks.

A key issue that leads to the vulnerability of the GCNs is the “message passing” scheme, in which the representation of each node is based on the aggregation of the embeddings of its neighbors\cite{ZWWSurvey}. As a result, the adversarial attacks could be propagated in the network and indirect attacks are feasible.

In order to increase the robustness, a natural solution is to control the propagation of messages among the nodes. Since edges are the foundation of message passing, a simple method is to recognize and delete adversarial edges during the preprocessing of the dataset\cite{Jaccard}. Another idea is to use attention mechanisms, which have been widely applied in both natural language processing models and graph neural networks\cite{RGCN, PAGNN}. Besides being able to process data with different input sizes, another benefit of attention mechanisms is that they provide a framework to control the message-passing processes according to the node features.

While being able to increase the robustness of the GCN models, the previous methods also have their drawbacks. Firstly, they are shown to be more effective under low perturbation rates. When the perturbation rates are high, these models are also vulnerable\cite{WJSurvey}. Secondly, some methods are unusable for datasets in which the node features are invisible\cite{Jaccard}. Finally, to the best of our knowledge, most defense methods are mainly designed for targeted attacks\cite{Jaccard, RGCN}.

In this paper, we propose Influence GCN (I-GCN), which takes advantage of a novel mechanism called influence mechanism. Similar to attention mechanisms, the influence mechanism can control the message passing process according to the node features. However, the influence mechanism offers a brand new perspective. During the message passing process, a node is able to exert two different types of influences. The introverted influence tries to maintain the embeddings of itself while the extroverted influence tries to influence the hidden representations of its neighbors.

We control the proportions of the two types of influence by a coefficient called Degree of Introverted Influence (DoII). For instance, The more the embedding of a node changes during the message passing, the more likely that its latent representation is influenced by the adversarial attack. In this situation, we assign lower DoII coefficient to the node while giving higher DoII scores to its neighbors to counteract the possible adversarial attack.

\begin{figure}[t]
\centering
\includegraphics[width=0.4\textwidth]{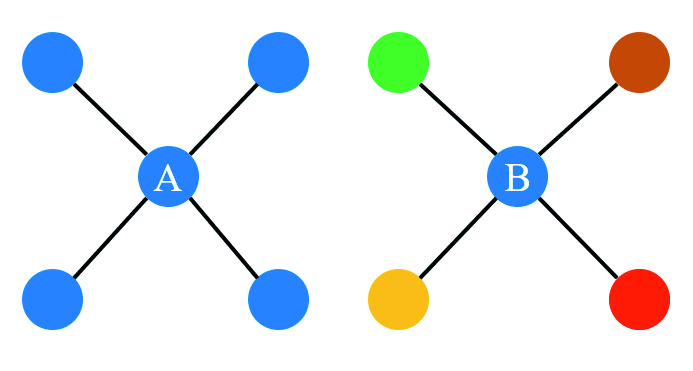}
\caption{An illustration of the fundamental idea of the influence mechanism.}
\label{fig1}
\end{figure}

In Figure \ref{fig1}, the node A is more similar to its neighbors than node B. Thus, node B is more likely to be attacked. As a result, node B is supposed to have a higher DoII coefficents. Extensive experiments have been conducted to verify the efficacy of our influence mechanism. We propose two variants of the influence mechanism and the experiments show that our I-GCN model outperforms the state-of-the-art methods under high perturbation rates.

In summary, our contributions of this paper are as follows:

\begin{itemize}
\item We propose the influence mechanism, which divides the effects of nodes into two different types, for the first time to the best of our knowledge.
\item We propose the I-GCN model, which is proved to outperform the state-of-the-art baselines when defending against non-targeted attacks.
\item We propose a variant of our model especially for datasets in which node features are not available.
\item We conduct extensive experiments to prove the efficacies of our proposed methods.
\end{itemize}

The rest of the paper is organized as follows. In Section 2, we review related works. In Section 3, we formally define the mathematical notations used in this paper and then introduce the preliminaries. Our proposed methods are explained in details in Section 4 while the experimental results are reported in Section 5. Finally, we conclude the paper in Section 6.

\section{Related Work}

Our work is based on recent researches on graph convolutional networks, graph adversarial attacks and graph adversarial defenses.

\subsection{Graph Convolutional Networks}
Graph Convolutional Networks (GCNs) are a kind of Graph Neural Networks (GNNs) which aims to generalize the convolution operation to the graph domain. According to the convolution operations defined in the networks, GCNs are divided into two categories:  spectral methods and spatial methods\cite{ZWWSurvey}. Bruna et al. first introduced the spectral method, which is based on spectral graph theory, using the graph Laplacian matrix\cite{Bruna}. Due to the eigen-decomposition operations of the graph Laplacians, the time complexities of such convolution methods are unsatisfactory. ChebNet\cite{Cheb} approximates the convolution operation with a K-order Chebyshev polynomial, thus improving the efficiency significantly. The GCN model\cite{Kipf} proposed by Kipf and Welling further simplify the convolution to first order and it achieves state-of-the-art performance on node classification tasks. We also choose GCN as an important baseline in our experiments.

However, spectral methods have natural weakness in generalizing among different graphs since the graph Laplacians used during the training are based upon certain graphs. Spatial methods define the graph convolution by aggregating information from the neighborhoods. Graph Attention Network (GAT)\cite{GAT} is a classic model in the spatial domain that learns the attention scores in an edgewise manner. Several general frameworks in the spatial domain such as GraphSAGE\cite{GraphSAGE} and MoNet\cite{MoNet} have been proposed. 

For an in-depth review of the GCNs, we refer the readers to recent surveys and texts\cite{ZWWSurvey, SSMSurvey, LiuIntro}.

\subsection{Graph Adversarial Attack}
While achieving state-of-the-art performances in different kinds of tasks, GCNs have been shown to be vulnerable to adversarial attacks recently\cite{WJSurvey}. Graph attack algorithms could be divided into different categories based on different standards. Especially, we can divide attacking algorithms into targeted attacks and non-targeted attack. Targeted attacks aim to misclassifying certain target nodes while non-targeted attacks focus on reducing the overall performance of the model of the full dataset\cite{ZSSurvey}. 

Nettack\cite{Nettack} is a targeted attack algorithm based on a greedy strategy which scores all possible perturbations and then chooses the one with the largest score in each step.  While also being a targeted algorithm, RL-S2V\cite{RLS2V} employs the technique of reinforcement learning to generate perturbations.

To attack the network globally, Mettack\cite{Mettack} is a poisoning non-targeted attack which poisons the edges of the graph before the training procedure. Being the first non-target attack algorithm, Mettack is also chosen as the attack algorithm in this paper.

\subsection{Graph Adversarial Defense}
In order to improve the robustness of the GCNs, several defense algorithms have been proposed by various researchers. Wu et al. found that targeted attack algorithms tend to connect nodes with large Jaccard similarity score and they propose the GCN-Jaccard model\cite{Jaccard} which remove dissimilar edges during the preprocessing of the dataset. Zhu et al. propose Robust GCN (RGCN) network\cite{RGCN} which adopts the Gaussian distributions of node features as the hidden representations. Believing that the nodes that are severely influenced by the attack would have higher variances, RGCN assigns low attention scores to them in order to defense against the attacker.

\section{Preliminaries}
Before explaining the details of the influence mechanism and our proposed I-GCN model, we introduce the notations used in this paper and some basic concepts.

\subsection{Notations}

In this paper we define a graph as $G=(V,E,X)$,  where $V=\{v_{1}, \cdots, v_{N}\}$ is the set of nodes, $N=|V|$ is the number of nodes, $E \subseteq V \times V$ is the set of edges, $X=\{x_{1}, \cdots, x_{N}\}$ is the set of node features. We denote the adjacency matrix of the graph as $A=[A_{1}, \cdots, A_{N}]\in \mathbb{R}^{N\times N}$ and the $N\times N$  identity matrix is denoted as $I_{N}$. For the set of nodes, we define three subsets. $V_{train} \subset V$  is the training set, $V_{val}\subset V$ is the validation set while $V_{test} \subset V$ is the testing set. Since both the baselines and our proposed model contain more than one layer, we use $H^{(l)}=[h_1^{(l)}, \cdots, h_N^{(l)}]$ to define the hidden representations of nodes in the $l^{th}$  layer of the model. Similarly, the weight matrix for the $l^{th}$ layer is denoted as $W^{(l)}$.

\subsection{Graph Convolutional Network}
We introduce the architecture of the GCN model. Although a number of graph convolutional networks have been proposed, in this paper we only consider the one proposed by Kipf and Welling\cite{Kipf}. In this model, a convolution layer aggregates and transforms the embeddings of all first-order neighbors for each node. The message passing mechanism could be described as follows:
\begin{equation}
H^{(l+1)} = \sigma(\tilde{D}^{-\frac{1}{2}}\tilde{A}\tilde{D}^{-\frac{1}{2}}H^{(l)}W^{(l)}),
\end{equation}
where $\tilde{A}=A+T_{N}$, $\tilde{D}_{ii} = \sum_{j} \tilde{A}_{ij}$ and $\sigma$ is an activation function such as ReLU. The trainable weight matrix $W^{(l)}$ performs the linear transformation that changes the dimensions of latent representations while $\hat{A} = \tilde{D}^{-\frac{1}{2}}\tilde{A}\tilde{D}^{-\frac{1}{2}}$ passes messages among the nodes according to the adjacency matrix.

The two-layer GCN network could be defined as:

\begin{equation}
f(X,A)=softmax(\hat{A}\sigma(\hat{A}XW^{(0)})W^{(1)}),
\end{equation}

For semi-supervised node classification tasks, the GCN model could be trained with the cross entropy loss function:
\begin{equation}
\mathcal{L}_{GCN}=-\sum_{i\in V_{train}} \sum_{j=1}^C Y_{if} \ln Z_{if}.
\end{equation}
where $C$ is the number of classes, $Y$ is the label matrix and $Z$ is the output of the network.

\section{The Proposed Methods}
In this section, we will introduce our influence mechanism and the Influence GCN (I-GCN) model. We first introduce the influence mechanism in details, and then we show the complete framework and the two variants of I-GCN.

\subsection{The Influence Mechanism}

\begin{figure*}[t]
\centering
\includegraphics[width=0.8\textwidth]{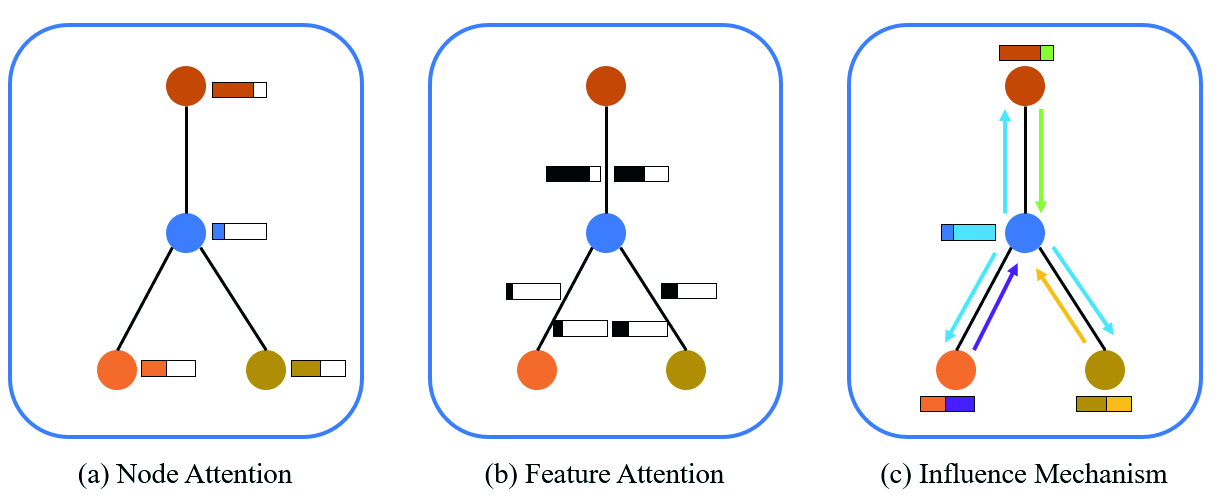}
\caption{A comparison between the influence mechanism and attention mechanisms. The colors of nodes represent the node feature. In (a) and (b), the bars reveal the attention scores while in (c), the bars represent the DoII coefficients and the effects of extroverted influence.}
\label{compare}
\end{figure*}

We will start by describing the general framework of the influence mechanism. The input to the layer is a set of hidden representations $H^{(l)}=[h_1^{(l)}, \cdots, h_N^{(l)}]\in\mathbb{R}^(N\times F_l)$ and the adjacency matrix  $A\in\mathbb{R}^{N\times N}$. In order to transform the hidden representations, a weight matrix $W^{(l)} \in \mathbb{R}^{F_l \times F_{l+1}}$, which performs the learnable linear transformation, is applied.

The influence layer assigns each node a DoII coefficient $\alpha_i^{(l)}$. We denote the DoII coefficients of all nodes as $\vec{\alpha}^{(l)}=[\alpha_1^{(l)},\cdots ,\alpha_N^{(l)}]$ and the DoII matrix is defined as:$M_{\alpha}^{(l)} = [\vec{\alpha}^{{(l)}^T},\cdots,\vec{\alpha}^{{(l)}^T}]^T$. Since the DoII coefficient indicates the degree of influence that a node exerts on its neighbors, we define the influenced adjacency matrix as:
\begin{equation}
A_I^{(l)} = \left( (I\odot M_{\alpha}^{(l)} + (M_1-I)\odot (M_1 - M_{\alpha}^{(l)} )\right) \odot \hat{A}
\label{Passing}
\end{equation}
where $M_1$ is the $N\times N$ matrix such that $\forall i,j \in [1,N], M_{1_{ij}} = 1$, $\odot$ denotes the Hadamard product operation and $\hat{A}$ is the normalized adjacency matrix which utilizes the renormalization trick\cite{Kipf}.

Figure \ref{compare} reveals the differences among node attentions, edge attentions and our proposed influence mechanism. In Figure \ref{compare}(c), the blue node is attacked that it is highly influenced by the brown, orange and yellow nodes. In order to counteract such attacking effects, on one hand, we increase the DoII coefficients of the 3 adversarial neighbors such that they will attack less during the message passing process. On the other hand, we decrease the DoII coefficient of the blue node such that it will attack the adversarial neighbors instead.

\subsection{The I-GCN Model}

\begin{figure}[t]
\centering
\includegraphics[width=0.4\textwidth]{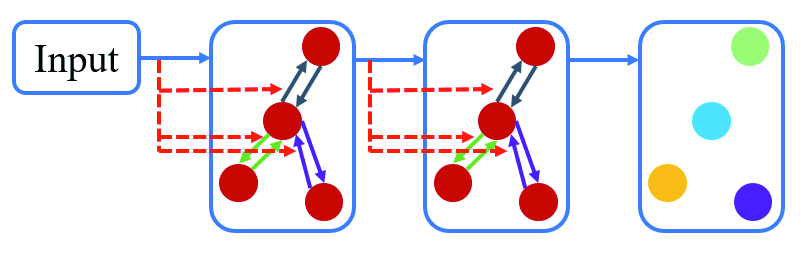}
\caption{The structure of I-GCN. The influence mechanism is illustrated in red dashed arrows.}
\label{igcn}
\end{figure}

In this subsection, we introduce our proposed I-GCN model in details. Similar to the GCN model proposed by Kipf and Welling\cite{Kipf}, our I-GCN model contains two graph influence layers. The general framework is illustrated in Figure \ref{igcn}. Both graph influence layers calculate DoII coefficients according to a DoII function for all nodes in the graph and then perform the message passing process. The output of the second graph influence layer is fed into a Softmax classifier which produces the final output of the I-GCN network. The I-GCN network could be defined as:

\begin{equation}
f(X,A)=Softmax\left(A_I^{(2)} \sigma(A_I^{(1)} X W^{(1)}) W^{(2)}\right).
\label{Network}
\end{equation}

The I-GCN model could be trained with the cross entropy loss functions:
\begin{equation}
\mathcal{L}_{I-GCN}=-\sum_{i\in V_{train}} \sum_{j=1}^C Y_{if} \ln f(X,A)_{if},
\label{Loss}
\end{equation}
where $C$ is the number of classes and $Y$ is the label matrix.

Then we introduce the detailed methods to assign DoII coefficients. Similar to attention mechanisms, a number of different methods to assign DoII scores could be deployed. Hence, our proposed I-GCN model is indeed a general framework with multiple variants. In this paper, we propose two variants of I-GCN which use different DoII functions to assign DoII coefficients. 

\begin{table*}[t]
\centering
\begin{tabular}{l | l c c c c}
Dataset & Type & Nodes & Edges & Classes & Features \\ \hline
Cora & Citation network & 2485 & 5069 & 7 & 1433\\
Citeseer & Citation network & 2110 & 3668 & 6 & 3703\\
Polblogs & Blog network & 1222 & 16714 & 2 & N/A\\
Cora-ML & Citation network & 2810 & 7981 & 7 & 2879 \\ \hline
\end{tabular}
\caption{Dataset statistics. We only consider nodes and edges in the largest connected components.}
\label{dataset}
\end{table*}

\subsubsection{I-GCN-N and the L1-Norm Based DoII Function}
The first strategy we choose to assign DoII coefficients is to consider the similarity of the hidden representations $h_i^{(l)}$ 
and  $h_i^{(l+1)'}$ 
where $h_{i}^{(l+1)'}=A_i h_i^{(l)}$ 
is a simulation of $h_i^{(l+1)}$ 
without applying the linear transformation $W^{(l)}$. Since higher similarity between $h_i^{(l)}$ 
and $h_i^{(l+1)}$ indicates fewer uncertainties and lower possibility for node i to be attacked, we assign different influence scores to the nodes as follows:
\begin{equation}
\alpha_i^{(l)} = \exp \left(-\lVert A_i h_i^{(l)} - h_i^{(l)}\rVert_1\right),
\label{Norm}
\end{equation}
in which the exponential function flips the monotonicity of the DoII scoring function. It also ensures that $\alpha_i \in (0,1]$.

Utilizing the L1-Norm based DoII function, we propose I-GCN-N, which is the first variant of I-GCN. Each layer of I-GCN-N assigns DoII functions according to (\ref{Norm}) and then propagates messages by equation (\ref{Passing}). The whole network propagates as (\ref{Network}).

\subsubsection{I-GCN-A and The Automatic DoII function}

While the L1-Norm based DoII function assigns DoII coefficients to the nodes according to the hidden representations, in some certain datasets node features are not visible. In such situations, it is expected that the hidden representations are not informative enough for scoring.

Thus, we propose the automatic DoII function that all the DoII coefficients are trained automatically by the optimizer. $N$ trainable parameters $\alpha_1, \cdots ,\alpha_N$ are created and they are used as the DoII coefficients.

Utilizing the automatic DoII function, we propose I-GCN-A, which is the second variant of I-GCN. Training DoII coefficients automatically, each layer of I-GCN-A passes messages according to equation (\ref{Passing}) directly and the whole network propagates as (\ref{Network}).

\section{Experiments}
\begin{table*}[t]
\centering
\begin{tabular}{c c | c c c c c}
\hline
Dataset & Ptb Rate & GCN & RGCN & GCN-Jaccard & I-GCN-N & I-GCN-A \\ \hline
\multirow{6}{*}{Cora} 
& 0 
& 83.27$\pm$0.79 
& \textbf{85.52$\pm$0.24} 
& 82.33$\pm$0.36 
& 84.27$\pm$0.37 
& 84.19$\pm$0.21 \\ 

& 0.05 
& 76.88$\pm$2.00 
& 78.92$\pm$1.74 
& 77.64$\pm$1.07 
& \textbf{79.12$\pm$1.4} 
& 79.31$\pm$1.18 \\

& 0.1 
& 71.11$\pm$1.29 
& 74.18$\pm$1.19 
& 74.72$\pm$1.31 
& \textbf{75.00$\pm$1.24} 
& 74.87$\pm$0.83 \\

& 0.15 
& 65.90$\pm$3.08 
& 69.03$\pm$2.08 
& \textbf{71.78$\pm$0.90} 
& 71.49$\pm$1.46 
& 71.23$\pm$1.48 \\

& 0.2 
& 59.45$\pm$2.29
& 63.81$\pm$2.06
& \textbf{68.29$\pm$1.70}
& 68.07$\pm$1.26
& 68.22$\pm$1.51 \\

& 0.25 
& 58.28$\pm$6.71
& 60.77$\pm$5.89
& \textbf{66.10$\pm$3.75}
& 65.19$\pm$4.58
& 65.49$\pm$4.86 \\ \hline

\multirow{6}{*}{Citeseer} 
& 0
& 76.34$\pm$0.31 
& \textbf{76.43$\pm$0.29}
& 73.80$\pm$0.94
& 74.29$\pm$0.32
& 74.01$\pm$0.56 \\

& 0.05
& \textbf{73.77$\pm$0.57}
& 73.55$\pm$0.34
& 70.55$\pm$1.26
& 72.36$\pm$0.55
& 72.35$\pm$0.53 \\

& 0.1
& \textbf{70.94$\pm$1.22}
& 70.50$\pm$0.53
& 66.50$\pm$1.67
& 70.47$\pm$1.56
& 70.35$\pm$1.07 \\

& 0.15
& 67.34$\pm$2.09
& 66.81$\pm$1.52
& 66.45$\pm$1.18
& \textbf{68.32$\pm$1.54}
& 68.07$\pm$1.41 \\

& 0.2
& 61.30$\pm$1.65
& 61.36$\pm$1.96
& 62.69$\pm$1.19
& \textbf{65.33$\pm$1.31}
& 64.98$\pm$1.31 \\

& 0.25
& 56.94$\pm$3.43 
& 57.70$\pm$2.20
& 61.00$\pm$2.39
& \textbf{63.45$\pm$2.99}
& 62.73$\pm$2.07 \\ \hline

\multirow{6}{*}{Polblogs}
& 0
& 95.43$\pm$0.32
& \textbf{95.62$\pm$0.14}
& N/A
& 93.93$\pm$0.91
& 92.78$\pm$2.42 \\

& 0.05
& 80.20$\pm$1.15 
& 80.06$\pm$1.11
& N/A
& 86.16$\pm$4.29
& \textbf{87.19$\pm$2.47} \\

& 0.1
& 79.40$\pm$1.79
& 79.51$\pm$1.08
& N/A
& 86.19$\pm$2.32
& \textbf{87.44$\pm$2.20} \\

& 0.15
& 79.49$\pm$1.37
& 79.39$\pm$1.04
& N/A
& 86.61$\pm$3.78
& \textbf{87.79$\pm$2.16} \\

& 0.2
& 79.12$\pm$1.52 
& 79.28$\pm$1.12
& N/A
& 86.84$\pm$1.40
& \textbf{87.59$\pm$2.11} \\

& 0.25
& 79.54$\pm$1.76
& 79.71$\pm$0.67
& N/A
& 85.03$\pm$2.83
& \textbf{87.12$\pm$1.70} \\ \hline

\multirow{6}{*}{Cora-ML}
& 0
& 83.96$\pm$1.27
& \textbf{86.32$\pm$0.29}
& 85.25$\pm$0.31
& 85.56$\pm$0.17
& 85.33$\pm$0.34 \\

& 0.05
& 79.72$\pm$1.74
& \textbf{84.31$\pm$0.36}
& 81.55$\pm$0.46
& 83.59$\pm$0.46
& 83.49$\pm$0.21 \\

& 0.1
& 68.01$\pm$4.26
& 80.08$\pm$0.56
& 76.86$\pm$0.70
& \textbf{82.07$\pm$0.40}
& 81.76$\pm$0.44 \\

& 0.15
& 58.57$\pm$3.11
& 75.20$\pm$0.47
& 73.45$\pm$0.81
& \textbf{80.06$\pm$0.71}
& 79.98$\pm$0.74 \\

& 0.2
& 49.92$\pm$4.54
& 68.98$\pm$0.80
& 71.14$\pm$0.93
& \textbf{78.01$\pm$0.73}
& 77.39$\pm$0.95 \\

& 0.25
& 46.06$\pm$3.13
& 64.12$\pm$1.33
& 67.84$\pm$0.95
& \textbf{76.36$\pm$1.00}
& 76.02$\pm$0.62 \\ \hline

\end{tabular}
\caption{Node classification performances (Accuracy$\pm$Std) under Mettack-LL. GCN-Jaccard is not applicable to the Polblogs dataset since the node features are not visible.}
\label{result}
\end{table*}
\begin{figure*}[t]
\centering
\includegraphics[width=1.0\textwidth]{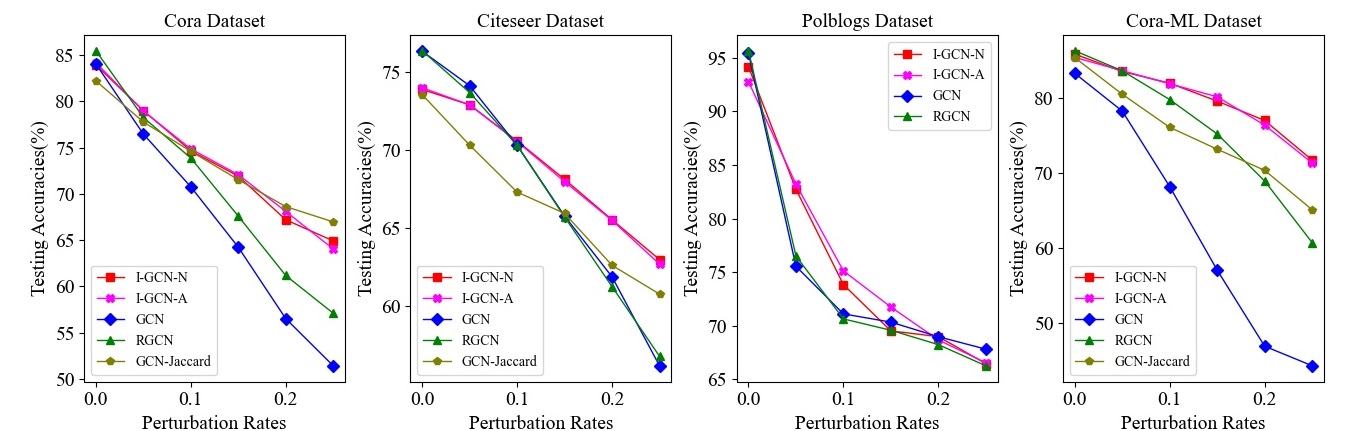}
\caption{Node classification accuracies under Mettack-F.}
\label{mettackf}
\end{figure*}

In this section, we evaluate the effectiveness of our proposed I-GCN model. We first introduce our experiment settings, datasets and baselines, then we show our experiment results.

\subsection{Experimental Settings}

\subsubsection{Datasets}

We validate our proposed methods on four different datasets that are commonly used in previous works \cite{Kipf, Mettack, RGCN}. In order to evaluate the proposed model comprehensively, we especially adopt the Polblogs dataset in which node features are not visible. The statistics of the datasets are summarized in Table \ref{dataset}.

Following \cite{Mettack}, we only consider the largest connected components of the graphs. We split the datasets randomly such that 10\% of the nodes are used for training, another 10\% of the nodes are used for validation and the remaining 80\% of the nodes form the testing set.

\subsubsection{Baselines}

To evaluate the robustness of I-GCN, we compare it with the most representative GCN model and two state-of-the-art defense algorithms.

\begin{itemize}
\item GCN\cite{Kipf}: As introduced in preliminaries, this is the most representative one among all the GCN models.
\item RGCN\cite{RGCN}: RGCN is a state-of-the-art defense model based on node attentions. As introduced in preliminaries, the model learns the hidden representations as Gaussian distributions and then assigns attention scores according to the variances.
\item GCN-Jaccard\cite{Jaccard}: GCN-Jaccard is a state-of-the-art defense algorithm based on pre-processing. Computing the Jaccard similarities of the features for all pairs of adjacent nodes, the algorithm deletes the edges with low Jaccard similarities since such edges are more likely to be adversarial edges.
\end{itemize}

\subsubsection{Adversarial Attacks}

The adversarial attack method we choose in this paper is the Mettack algorithm\cite{Mettack}. Mettack is the first and state-of-the-art non-targeted attacking algorithm which attacks graphs via meta-gradients. In this paper, we utilize the Mettack algorithm implemented in the DeepRobust library\cite{DeepRobust}. We apply two variants of Mettack algorithm. The Mettack-LL variant enforces the degree distribution constraint while the Mettack-F variant frees this constraint. The perturbation rate is varied from 0\% to 25\% with a step of 5\%.

\subsubsection{Parameter Settings}

For GCN, GAT and RGCN, we adopts the hyper-parameters  used in authors' implementations. For I-GCN-N, we train the model for 200 epochs with an early stopping patience of 20 epochs. We adopt the early stopping method of the GAT\cite{GAT} model. Varying the number of hidden units from $\{8, 16, 24, 32, 40, 48\}$, we choose 40 as the number of hidden unit. Dropout rate is 0.5. The activation function is ReLU. For the optimization, we use the Adam optimizer with fixed learning rate of 0.01. The I-GCN-A model uses the same parameter settings and all the trainable DoII values are initialized as 0.

\subsubsection{Experimental Environment}

Our experiments are conducted on a server with the following configurations:

\begin{itemize}
\item CPU: Intel Core i9-10980XE
\item GPU: Nvidia Quadro RTX 6000 with CUDA 10.2
\item Operating System: Microsoft Windows 10
\item Python: Python 3.7.6
\item Libraries: PyTorch 1.5.0, Numpy 1.18.5, Scipy 1.3.1
\end{itemize}

\subsection{Experimental Results}

All the experiments are conducted 10 times. We report the average accuracies and standard deviations when adopting Mettack-LL in Table \ref{result}. The table shows that our models outperform other methods in most situations. The accuracies of different methods under Mettack-F are reported in Figure \ref{mettackf}. The following observations are concluded from the experimental results.

\begin{figure}[t]
\centering
\includegraphics[width=0.4\textwidth]{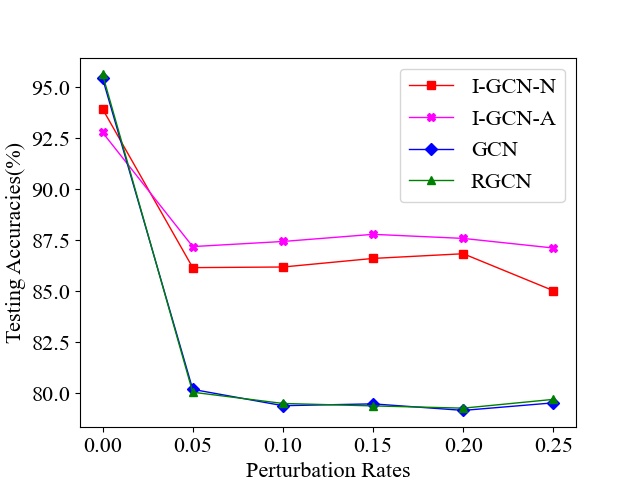}
\caption{Results of different methods on Polblogs dataset when adopting Mettack-LL.}
\label{polblogs}
\end{figure}

\begin{itemize}
\item Both variants of our proposed I-GCN model outperform other methods consistently under high perturbation rates for most datasets. When defending against Mettack-LL on Cora-ML dataset, our I-GCN-N model improves original GCN, RGCN and GCN-Jaccard by 30\%, 12\% and 8\% respectively under the 25\% perturbation rate.
\item While performing well on Cora dataset, GCN-Jaccard performs worse than our proposed I-GCN model on other datasets. Meanwhile, our model outperforms all other baselines on Cora. This shows the adaptability of our I-GCN model.
\item For the Polblogs dataset, GCN-Jaccard is not applicable since node features are not available. Although being applicable, RGCN fails to guarantee any significant improvements on the original GCN. When defending against Mettack-LL, our proposed I-GCN-A model improves 9\% under the 25\% perturbation rate. When defending against Mettack-F, our proposed I-GCN-A model improves 10\% under the 5\% perturbation rate. A comparison of different methods on Polblogs is shown in Figure \ref{polblogs}.
\end{itemize}

\subsection{Parameter Analysis}
In this subsection, we conduct some parameter analysis to investigate the sensitivity of the hyperparameters. For fair comparison, we use the same learning rate and weight decay as the original GCN paper. 

\begin{figure}[t]
\centering
\includegraphics[width=0.4\textwidth]{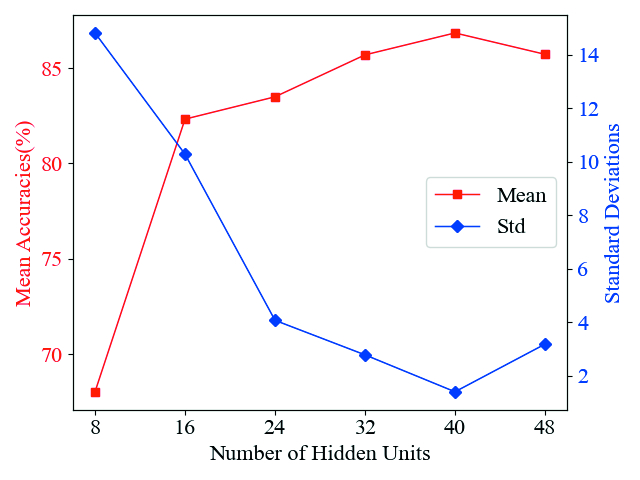}
\caption{Performances of I-GCN-N when defending against Mettack-LL on Polblogs dataset with perturbation rate 0.2.}
\label{hidden}
\end{figure}

We alter the number of hidden units among $\{8, 16, 24, 32, 40, 48\}$. For the convenience of presentation, we use Polblog dataset with 20\% perturbation rate as an example. Figure \ref{hidden} shows the performance of I-GCN-N when defending against Mettack-LL. We can see that a low number of hidden units will decrease the stability of the I-GCN-N network. Both the mean and the standard deviation of the accuracy are relatively smooth when the number of hidden units varies on the interval $[24, 48]$.

\section{Conclusion}

While achieving state-of-the-art performances in the task of node classifications, GCNs can be easily fooled by adversarial perturbations. Several defense algorithms have been proposed but most of them mainly aim to defend against targeted attacks such as Nettack. Meanwhile, experiments show that when the graphs are heavily poisoned, the performances of the defense methods also reduce significantly. In this paper, we propose a novel I-GCN model which utilizes a new mechanism called influence mechanism. Experimental results show the effectiveness of our proposed model under high perturbation rates.

For datasets without node features, we propose the I-GCN-A variant of our model such that all the coefficients are trained automatically. Our experiments on Polblogs dataset demonstrate the robustness of our model on such datasets. 

Future directions include developing more DoII functions and conducting more experiments to reveal the effects of the influence mechanism. We also aim to extend this framework to other graph-based deep learning models such as multi-graph networks.

\bibliography{test.bib}

\begin{thebibliography}{23}
\providecommand{\natexlab}[1]{#1}
\providecommand{\url}[1]{\texttt{#1}}
\providecommand{\urlprefix}{URL }
\expandafter\ifx\csname urlstyle\endcsname\relax
  \providecommand{\doi}[1]{doi:\discretionary{}{}{}#1}\else
  \providecommand{\doi}{doi:\discretionary{}{}{}\begingroup
  \urlstyle{rm}\Url}\fi

\bibitem[{Bruna et~al.(2014)Bruna, Zaremba, Szlam, and LeCun}]{Bruna}
Bruna, J.; Zaremba, W.; Szlam, A.; and LeCun, Y. 2014.
\newblock Spectral Networks and Locally Connected Networks on Graphs.
\newblock In \emph{2nd International Conference on Learning Representations,
  {ICLR} 2014, Banff, AB, Canada, April 14-16, 2014, Conference Track
  Proceedings}.

\bibitem[{Chen et~al.(2020)Chen, Li, Peng, Xie, Cao, Xu, He, and
  Zheng}]{ZSSurvey}
Chen, L.; Li, J.; Peng, J.; Xie, T.; Cao, Z.; Xu, K.; He, X.; and Zheng, Z.
  2020.
\newblock A Survey of Adversarial Learning on Graphs.
\newblock \emph{CoRR} abs/2003.05730.

\bibitem[{Dai et~al.(2018)Dai, Li, Tian, Huang, Wang, Zhu, and Song}]{RLS2V}
Dai, H.; Li, H.; Tian, T.; Huang, X.; Wang, L.; Zhu, J.; and Song, L. 2018.
\newblock Adversarial Attack on Graph Structured Data.
\newblock In \emph{Proceedings of the 35th International Conference on Machine
  Learning, {ICML} 2018, Stockholmsm{\"{a}}ssan, Stockholm, Sweden, July 10-15,
  2018}, volume~80 of \emph{Proceedings of Machine Learning Research},
  1123--1132. {PMLR}.

\bibitem[{Defferrard, Bresson, and Vandergheynst(2016)}]{Cheb}
Defferrard, M.; Bresson, X.; and Vandergheynst, P. 2016.
\newblock Convolutional Neural Networks on Graphs with Fast Localized Spectral
  Filtering.
\newblock In \emph{Advances in Neural Information Processing Systems 29: Annual
  Conference on Neural Information Processing Systems 2016, December 5-10,
  2016, Barcelona, Spain}, 3837--3845.

\bibitem[{Hamilton, Ying, and Leskovec(2017)}]{GraphSAGE}
Hamilton, W.~L.; Ying, Z.; and Leskovec, J. 2017.
\newblock Inductive Representation Learning on Large Graphs.
\newblock In \emph{Advances in Neural Information Processing Systems 30: Annual
  Conference on Neural Information Processing Systems 2017, 4-9 December 2017,
  Long Beach, CA, {USA}}, 1024--1034.

\bibitem[{Jin et~al.(2020)Jin, Li, Xu, Wang, and Tang}]{WJSurvey}
Jin, W.; Li, Y.; Xu, H.; Wang, Y.; and Tang, J. 2020.
\newblock Adversarial Attacks and Defenses on Graphs: {A} Review and Empirical
  Study.
\newblock \emph{CoRR} abs/2003.00653.

\bibitem[{Kipf and Welling(2017)}]{Kipf}
Kipf, T.~N.; and Welling, M. 2017.
\newblock Semi-Supervised Classification with Graph Convolutional Networks.
\newblock In \emph{5th International Conference on Learning Representations,
  {ICLR} 2017, Toulon, France, April 24-26, 2017, Conference Track
  Proceedings}.

\bibitem[{Li et~al.(2020)Li, Jin, Xu, and Tang}]{DeepRobust}
Li, Y.; Jin, W.; Xu, H.; and Tang, J. 2020.
\newblock DeepRobust: A PyTorch Library for Adversarial Attacks and Defenses.

\bibitem[{{Liu} and {Zhou}(2020)}]{LiuIntro}
{Liu}, Z.; and {Zhou}, J. 2020.
\newblock \emph{Introduction to Graph Neural Networks}.
\newblock Morgan and Claypool.

\bibitem[{Monti et~al.(2017)Monti, Boscaini, Masci, Rodol{\`{a}}, Svoboda, and
  Bronstein}]{MoNet}
Monti, F.; Boscaini, D.; Masci, J.; Rodol{\`{a}}, E.; Svoboda, J.; and
  Bronstein, M.~M. 2017.
\newblock Geometric Deep Learning on Graphs and Manifolds Using Mixture Model
  CNNs.
\newblock In \emph{2017 {IEEE} Conference on Computer Vision and Pattern
  Recognition, {CVPR} 2017, Honolulu, HI, USA, July 21-26, 2017}, 5425--5434.
  {IEEE} Computer Society.

\bibitem[{{Newman}, {Watts}, and {Strogatz}(2002)}]{Social}
{Newman}, M. E.~J.; {Watts}, D.~J.; and {Strogatz}, S.~H. 2002.
\newblock Random graph models of social networks.
\newblock \emph{Proceedings of the National Academy of Sciences of the United
  States of America} 99(90001): 2566--2572.

\bibitem[{Ron and Shamir(2013)}]{Transaction}
Ron, D.; and Shamir, A. 2013.
\newblock Quantitative Analysis of the Full Bitcoin Transaction Graph.
\newblock In \emph{Financial Cryptography and Data Security - 17th
  International Conference, {FC} 2013, Okinawa, Japan, April 1-5, 2013, Revised
  Selected Papers}, volume 7859 of \emph{Lecture Notes in Computer Science},
  6--24. Springer.

\bibitem[{{Rudolph} and {Cox}(2019)}]{Protein}
{Rudolph}, J.~D.; and {Cox}, J. 2019.
\newblock A Network Module for the Perseus Software for Computational
  Proteomics Facilitates Proteome Interaction Graph Analysis.
\newblock \emph{Journal of Proteome Research} 18(5): 2052--2064.

\bibitem[{Sen et~al.(2008)Sen, Namata, Bilgic, Getoor, Gallagher, and
  Eliassi{-}Rad}]{Citation}
Sen, P.; Namata, G.; Bilgic, M.; Getoor, L.; Gallagher, B.; and Eliassi{-}Rad,
  T. 2008.
\newblock Collective Classification in Network Data.
\newblock \emph{{AI} Mag.} 29(3): 93--106.

\bibitem[{Tang et~al.(2020)Tang, Li, Sun, Yao, Mitra, and Wang}]{PAGNN}
Tang, X.; Li, Y.; Sun, Y.; Yao, H.; Mitra, P.; and Wang, S. 2020.
\newblock Transferring Robustness for Graph Neural Network Against Poisoning
  Attacks.
\newblock In \emph{{WSDM} '20: The Thirteenth {ACM} International Conference on
  Web Search and Data Mining, Houston, TX, USA, February 3-7, 2020}, 600--608.
  {ACM}.

\bibitem[{Velickovic et~al.(2018)Velickovic, Cucurull, Casanova, Romero,
  Li{\`{o}}, and Bengio}]{GAT}
Velickovic, P.; Cucurull, G.; Casanova, A.; Romero, A.; Li{\`{o}}, P.; and
  Bengio, Y. 2018.
\newblock Graph Attention Networks.
\newblock In \emph{6th International Conference on Learning Representations,
  {ICLR} 2018, Vancouver, BC, Canada, April 30 - May 3, 2018, Conference Track
  Proceedings}.

\bibitem[{Wu et~al.(2019)Wu, Wang, Tyshetskiy, Docherty, Lu, and Zhu}]{Jaccard}
Wu, H.; Wang, C.; Tyshetskiy, Y.; Docherty, A.; Lu, K.; and Zhu, L. 2019.
\newblock Adversarial Examples for Graph Data: Deep Insights into Attack and
  Defense.
\newblock In \emph{Proceedings of the Twenty-Eighth International Joint
  Conference on Artificial Intelligence, {IJCAI} 2019, Macao, China, August
  10-16, 2019}, 4816--4823. ijcai.org.

\bibitem[{Yuan et~al.(2019)Yuan, He, Zhu, and Li}]{AdvI}
Yuan, X.; He, P.; Zhu, Q.; and Li, X. 2019.
\newblock Adversarial Examples: Attacks and Defenses for Deep Learning.
\newblock \emph{{IEEE} Trans. Neural Networks Learn. Syst.} 30(9): 2805--2824.

\bibitem[{Zhang, Cui, and Zhu(2018)}]{ZWWSurvey}
Zhang, Z.; Cui, P.; and Zhu, W. 2018.
\newblock Deep Learning on Graphs: {A} Survey.
\newblock \emph{CoRR} abs/1812.04202.

\bibitem[{Zhou et~al.(2018)Zhou, Cui, Zhang, Yang, Liu, and Sun}]{SSMSurvey}
Zhou, J.; Cui, G.; Zhang, Z.; Yang, C.; Liu, Z.; and Sun, M. 2018.
\newblock Graph Neural Networks: {A} Review of Methods and Applications.
\newblock \emph{CoRR} abs/1812.08434.

\bibitem[{Zhu et~al.(2019)Zhu, Zhang, Cui, and Zhu}]{RGCN}
Zhu, D.; Zhang, Z.; Cui, P.; and Zhu, W. 2019.
\newblock Robust Graph Convolutional Networks Against Adversarial Attacks.
\newblock In \emph{Proceedings of the 25th {ACM} {SIGKDD} International
  Conference on Knowledge Discovery {\&} Data Mining, {KDD} 2019, Anchorage,
  AK, USA, August 4-8, 2019}, 1399--1407. {ACM}.

\bibitem[{Z{\"{u}}gner, Akbarnejad, and G{\"{u}}nnemann(2018)}]{Nettack}
Z{\"{u}}gner, D.; Akbarnejad, A.; and G{\"{u}}nnemann, S. 2018.
\newblock Adversarial Attacks on Neural Networks for Graph Data.
\newblock In \emph{Proceedings of the 24th {ACM} {SIGKDD} International
  Conference on Knowledge Discovery {\&} Data Mining, {KDD} 2018, London, UK,
  August 19-23, 2018}, 2847--2856. {ACM}.

\bibitem[{Z{\"{u}}gner and G{\"{u}}nnemann(2019)}]{Mettack}
Z{\"{u}}gner, D.; and G{\"{u}}nnemann, S. 2019.
\newblock Adversarial Attacks on Graph Neural Networks via Meta Learning.
\newblock In \emph{7th International Conference on Learning Representations,
  {ICLR} 2019, New Orleans, LA, USA, May 6-9, 2019}.

\end{thebibliography}

\end{document}